%% file: main.tex
\documentclass[letterpaper, 10 pt, journal, twoside]{IEEEtran}  %
\IEEEoverridecommandlockouts

\usepackage{subfig}
\usepackage{xcolor}
\usepackage{lipsum}
\usepackage{mathtools}
\usepackage{cuted}
\usepackage{caption}
\usepackage{graphics}
\usepackage{epsfig} 
\usepackage{times} 
\usepackage{amsmath} 

\usepackage{amsthm}
\usepackage{amssymb}  

\usepackage{psfrag}
\usepackage{cancel}
\usepackage{tikz}
\usepackage{accents}
\usepackage{comment}
\usepackage{color}
\usepackage{array}
\usepackage{breqn}
\usepackage{multirow}
\usepackage{algorithm}
\usepackage{algpseudocode}

\theoremstyle{remark}
\newtheorem{remark}{Remark}
\renewcommand\vec\boldsymbol

\usepackage{caption}
\usepackage{booktabs}                     
\usepackage{tabularx}
\include{000_preamble}

\newcommand\blfootnote[1]{
\begingroup
\renewcommand\thefootnote{}\footnote{#1}
\addtocounter{footnote}{-1}
\endgroup
}

\begin{document}


 \markboth{IEEE ROBOTICS AND AUTOMATION LETTERS. PREPRINT VERSION. JUNE, 2022}%
 {Zhai \MakeLowercase{\textit{et al.}}: DA$^2$ Dataset: Toward Dexterity-Aware Dual-Arm Grasping}

\title{DA$^2$ Dataset: \\Toward Dexterity-Aware Dual-Arm Grasping}

\author{Guangyao Zhai$^{1}$, Yu Zheng$^{2}$,~\IEEEmembership{Senior Member,~IEEE}, Ziwei Xu$^{2}$, Xin Kong$^{3}$, Yong Liu$^{4}$,\\ Benjamin Busam$^{1}$, Yi Ren$^{2}$, Nassir Navab$^{1,5}$,~\IEEEmembership{Fellow,~IEEE},  and Zhengyou Zhang$^{2}$,~\IEEEmembership{Fellow,~IEEE}%
}

\twocolumn[{
\renewcommand\twocolumn[1][]{#1}
\maketitle
\begin{center}
\vspace{-3mm}
    \centering
    \captionsetup{type=figure}
    \includegraphics[width=\textwidth]{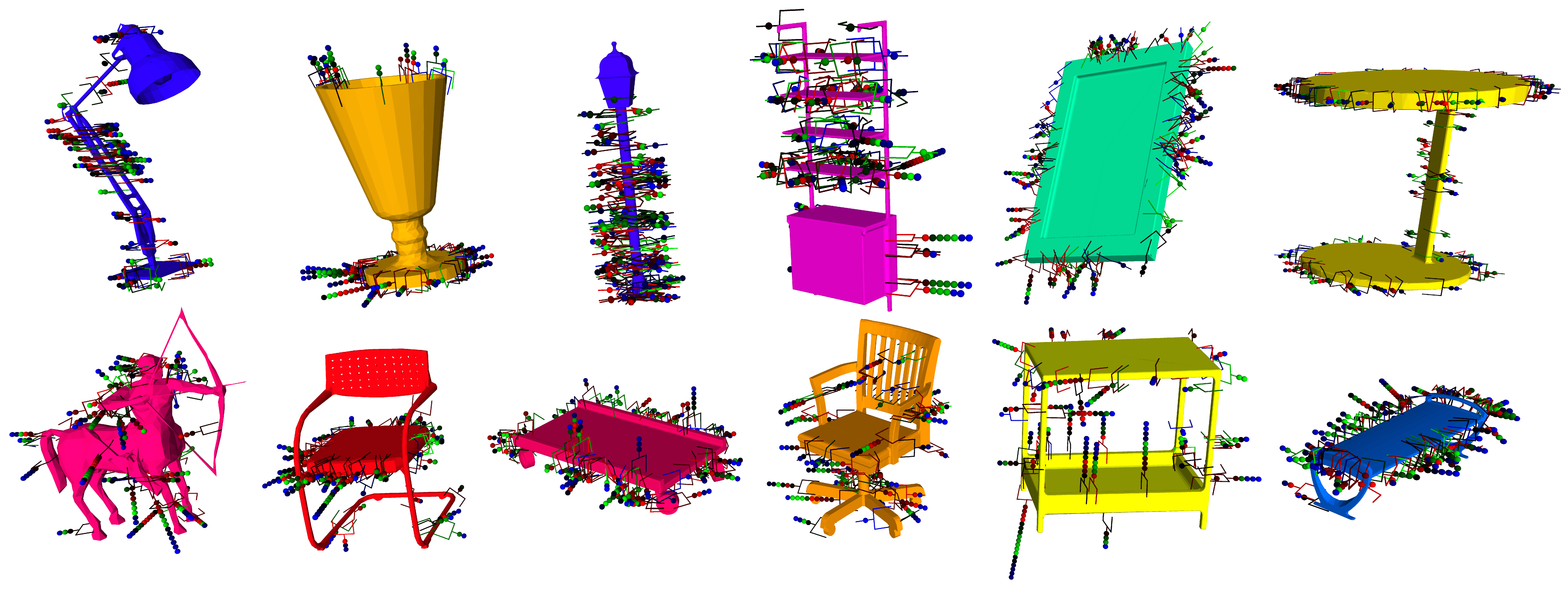}
    \captionof{figure}{A glimpse of the DA$^2$ dataset. Various grasp pairs are colored differently (best viewed with zoom-in). As one single grasp can be affiliated with several grasp pairs, we attach several sphere markers to the grasp to distinguish different pairs. The number of spheres on one grasp indicates how many times the grasp has been reused.}\label{intro}
    \vspace{-3mm}
\end{center}
}]


\blfootnote{
{Manuscript received: February 24, 2022; Revised: May 21, 2022; Accepted: June 27, 2022. This paper was recommended for publication by Editor Markus Vincze upon evaluation of the Associate Editor and Reviewers’ comments. This work was mainly done while Guangyao Zhai was an intern at Tencent Robotics X. \textit{(Corresponding author: Yi Ren)}}

{$^{1}$ Guangyao Zhai, Benjamin Busam and Nassir Navab are with the Chair for Computer  Aided Medical Procedures and Augmented Reality, Technical University of Munich, Munich, Germany {(guangyao.zhai@tum.de; b.busam@tum.de; nassir.navab@tum.de).}}

{$^{2}$ Yu Zheng, Ziwei Xu, Yi Ren and Zhengyou Zhang are with Tencent Robotics X, Tencent, Shenzhen, China {(petezheng@tencent.com; ziwei.xu@tum.de; evanyren@tencent.com; zhengyou@tencent.com).}}

{$^{3}$ Xin Kong is with the Department of Computing, Imperial College London, United Kingdom {(xk221@ic.ac.uk).}}

{$^{4}$ Yong Liu is with the Institute of Cyber-Systems and Control, Zhejiang University, Hangzhou, China {({yongliu@iipc.zju.edu.cn}).}}

{$^{5}$ Nassir Navab is also with the Laboratory for Computational Sensing and Robotics, Johns Hopkins University, Baltimore, MD, USA.}

{Digital Object Identifier (DOI): see top of this page.}
}

\begin{abstract}
In this paper, we introduce DA$^2$, the first large-scale dual-arm dexterity-aware dataset {for the generation of optimal bimanual grasping pairs for arbitrary large objects}. The dataset contains about {9M} pairs of parallel-jaw grasps, generated from {more than 6000}~objects and each labeled with various grasp dexterity measures. In addition, we propose an end-to-end dual-arm grasp
evaluation model trained on the rendered scenes from this dataset. We utilize the evaluation model as our baseline to show the value of this novel and nontrivial dataset by both online analysis and real robot experiments. All data and related code will be open-sourced at \url{https://sites.google.com/view/da2dataset}.
\end{abstract}


\IEEEpeerreviewmaketitle
\input{00_introduction.tex}

\input{01_related_work.tex}

\input{02_dataset.tex}

\input{04_graspnet.tex}
\input{05_results.tex}
\input{06_conclusion.tex}


\renewcommand*{\bibfont}{\footnotesize}
\printbibliography

\end{document}

%% file: 000_preamble.tex
\usepackage{hyperref}
\usepackage[backend=bibtex,
            hyperref=true,
            url=false,
            isbn=false,
            doi=false,
            backref=false,
            style=ieee,
            natbib=true,%
            mincitenames=1,
            maxcitenames=1,
            citestyle=numeric-comp,
            sorting=nyt,%
            block=none]{biblatex}
\usepackage{siunitx}

\usepackage{manfnt}
\usepackage{amssymb}
\usepackage{bbding}
\usepackage{fontawesome}
\usepackage{dsfont}
\usepackage{threeparttable}

\usepackage{makecell}
\usepackage{color, colortbl}
\newcolumntype{R}[2]{%
    >{\adjustbox{angle=#1,lap=\width-(#2)}\bgroup}%
    l%
    <{\egroup}%
}

\definecolor{LightGray}{gray}{0.9}

\definecolor{ben}{rgb}{0.9,0.,0.5}

\definecolor{similar}{rgb}{0.4,0.1,0.9}

%% file: 00_introduction.tex
\section{Introduction}
\IEEEPARstart{O}{ver} the past few years, numerous methods have surged to solve the problem of grasping arbitrary objects in a data-driven manner. The data sources being used significantly affect the success rate of these methods. As a result, many grasp datasets have been introduced~\cite{jiang2011efficient,zhang2019roi,levine2018learning,goldfeder2009columbia,morrison2020egad,chu2018real} to stipulate academic advancement. Although the research of single-arm grasping dealing with small objects has been well explored~\cite{lenz2015deep,levine2018learning,mahler2017dex,ten2017grasp,tobin2018domain, wang2021graspness}, how to address the query of dual-arm optimal grasping {for objects with various shapes} remains untouched. Dual-arm manipulation has some obvious advantages. For example, a dual-arm-hand (DAH) system can coordinately manipulate large and heavy objects with high dexterity which single-arm robots cannot accomplish. However, some problems in the robotic manipulation area may result in failures in exploiting these advantages by directly using current state-of-the-art approaches in terms of optimal grasp planning.

Unlike common tiny items, larger objects tend to be more sensitive to the positions a robot grasps since they usually have more complex geometric features and load distribution. This particularity results in the inapplicability of the current single-arm approaches to the dual-arm cases. To pave the way towards closing this gap, it is necessary to generate and select more feasible and reasonable grasp pairs for dual-arm cooperation and manipulation from the astronomical number of pairwise grasp configurations obtained by random combination. While the multi-fingered manipulation theories, including the dual-arm one~\cite{vahrenkamp2011bimanual}, have been well-developed over a long period, how to bridge the perception and classical dexterous grasping theory to facilitate the practical implementation of bimanual grasp is still unclear. The focus of research in the dual-arm grasping estimation field will narrow the gap between vision and manipulation, thus considerably enlarging application scenarios of the robots both in domestic and industrial environments beyond picking tiny objects from bins using a single arm.

To facilitate this process, a standardized grasping dataset related to DAH grasping is required. In the current single-arm grasping research area, fundamental methodologies for data generation can be classified into three categories: trial-and-error with grasping experiments~\cite{pinto2016supersizing,levine2018learning}, manual annotation~\cite{kappler2015leveraging,yan2018learning}, and synthetic data generation. The last category is subsequently classified into analytic grasp~\cite{mahler2017dex} and physics engine based simulation~\cite{kappler2015leveraging,depierre2018jacquard,eppner2021acronym}. Inspired by the pioneering work~\cite{eppner2021acronym}, who generate a grasp dataset from meshes with high flexibility, we also choose ShapeNet~\cite{savva2015semantically} to be our object data source. We generate dual-arm grasp poses labeled with various dexterity features. Our synthetic dataset combines several desirable properties reflecting multiple characteristics in dual-arm parallel-jaw grasping: 
\\\textbf{Various dexterity measures:}
Popular 3D-based methods~\cite{mahler2017dex,liang2019pointnetgpd,ten2017grasp} proposed in recent years process 3D estimation perfectly. Nevertheless, they are trained on data generated by considering force closure as the main factor influencing grasp performance. However, grasp dexterity analysis is also an important factor to evaluate the grasp candidate and is missing in the training process. Our dataset fully utilizes the grasp matrix and analyzes it from various aspects such as grasp stability and dexterity. Moreover, we design a novel evaluation module named Dual-PointNetGPD which is based on PointNetGPD~\cite{liang2019pointnetgpd} to select the optimal grasp pair from massive pair candidates through scoring of the visual inputs.
\\\textbf{Multiple applications:} Our dataset is compatible with common tools, like trimesh\footnote{\url{https://github.com/mikedh/trimesh}} and pyrender\footnote{\url{https://github.com/mmatl/pyrender}} to process collision checks and cluttered scene rendering. With a customized modification of the dataset, it can further serve for various manipulation tasks, especially for large objects eventually improving the autonomy of robotic systems for e.g. Robot-to-Human (R2H) handover, Human-robot collaboration, and autonomous bimanual manipulation.

The contributions of this work are as follows:
\begin{enumerate}
    \item We build a large-scale \textbf{D}ual-\textbf{A}rm \textbf{D}exterity-\textbf{A}ware (\textbf{DA$^2$}) grasp dataset, including a total of about {9M} parallel-jaw grasp pairs for {more than 6000} different meshes (see Fig.~\ref{intro}). The grasp pairs are labeled with multiple grasp dexterity measures by fully analyzing the grasp matrix. Our unique dataset constitutes a standardized data source filling the gap in vision-guided dual-arm grasping {of arbitrary objects}.
    \item We propose a novel grasp quality scoring module Dual-PointNetGPD as a baseline exclusive for DAH systems by modifying feature representations and coordinate expression of the single-arm method PointNetGPD.
    \item Both our online validation and real robot experiments verify that the combined measures introduced in our proposed dataset are practical and are easily learned by our baseline module through feature embedding of the rendered point clouds.
    
\end{enumerate}


%% file: 01_related_work.tex
\begin{table*}[!thbp]
\caption{Comparison of publicly available grasp datasets. }
    \begin{center}
        \begin{tabular}{lccrrccl}
    \toprule
        \multirow{2}{*}{Dataset} &  Grasps &  Grasp &  Total &  Total &  Data & Metric &  Gripper\\
         & /obj. & label & obj. & grasps & source & variety(ies) & support\\
    \midrule
        Levine et al.~\cite{levine2018learning} & N/A & Rect. & - & 800k & Real & 1 & Single \\
        Dex-Net~\cite{mahler2017dex} & 100 & 6-DOF & 1500 & 6.7M & Sim. & 1 & Single$^*$ \\
        GraspNet-1B~\cite{fang2020graspnet} & 12.5M & 6-DOF & 88 & 1.2B & Sim.+Real & 2 & Single \\
        PointNetGPD~\cite{liang2019pointnetgpd} & N/A & 6-DOF & 47 & 350k & Sim.+Real & 2 & Single \\
        ACRONYM\cite{eppner2021acronym} & 2000 & 6-DOF & 8872 & 17.7M & Sim. & 1 & Single \\
        DA$^2$ (Ours)  & up to 2001 & 6-DOF & 6327 & 9.0M & Sim. & \textbf{3} & \textbf{Dual} \\
    \bottomrule
    \end{tabular}
    \begin{threeparttable}
    \begin{tablenotes}
    \centering
    $^*$Dex-Net series support two-modality end-effectors which constitute a parallel jaw and suction cup, but they are not tightly coupled.
    \end{tablenotes}
    \end{threeparttable}
    \end{center}
    
    \label{tab:datasets}
    \vspace{-5mm}
\end{table*}

\section{Related Work}
\label{sec:related_work}
This section includes a review of some single-arm grasp datasets with related estimation methods. As our dataset is the first attempt to lay the basis for optimal dual-arm grasp pose generation, the general pipeline and evaluation baseline for the single-arm grasp pose generation are reviewed and analyzed here. Existing grasp datasets mostly focus on the single-arm-based grasp of small objects but differ in the type of data sources, labeling methods for the grasp samples as well as the intra-dataset variations (see Table.~\ref{tab:datasets}). It should be noted that only our dataset supports dual-arm grasping.

Robot grasping is an inherent 3D spatial manipulation task. Therefore, grasp configurations in \textit{SE}(3) show benefits compared with 2D planar representations~\cite{redmon2015real,wang2016robot,asif2017rgb,guo2017hybrid,kumra2017robotic,asif2018graspnet,chu2018real,zhou2018fully}. Similar to popular datasets for single-arm grasping~\cite{goldfeder2009columbia,kappler2015leveraging,mahler2017dex,veres2017integrated}, our dataset dispenses grasp pairs in 6-DOF format. 
Previous pioneering works provide real RGB-D measurements, however, are either limited in scale~\cite{jiang2011efficient} or limited to planar grasping~\cite{levine2018learning,zhang2019roi}. As described in ACRONYM~\cite{eppner2021acronym}, simulating observations provides a scalable alternative. Synthetic datasets are flexible and can be customized by incorporating either RGB images~\cite{depierre2018jacquard} or other modalities such as depth maps and point clouds. Some of them, e.g., Columbia grasp database~\cite{goldfeder2009columbia} and ACRONYM, can be seamlessly combined with any commonly-used renderers.
One of the differences among various datasets is the way the grasping qualities are labeled. While physics-based simulators~\cite{kappler2015leveraging,veres2017integrated,depierre2018jacquard} can be used for dataset creation, others leverage explicit models with analytic design~\cite{goldfeder2009columbia,mahler2017dex,fang2020graspnet} or practical exploration with real robots~\cite{levine2018learning}. Grasp data, however, can be cumbersome to annotate by hand~\cite{jiang2011efficient,kappler2015leveraging}.
Most methods~\cite{fang2020graspnet,mahler2017dex,liang2019pointnetgpd} that use analytical models leverage parallel-jaw force closure based on point contact with a friction model (PCWF). To comprehensively evaluate the grasp quality, a combined metric with Grasp Wrench Space (GWS) analysis and quantified force-closure inspection is employed in PointNetGPD~\cite{liang2019pointnetgpd}. The baseline in GraspNet-1B~\cite{fang2020graspnet} defines a measure called Grasp Affinity Fields (GAFs) to describe grasping robustness. To deal with dual-arm grasping, the single use of a stability descriptor is not sufficient. The challenge lies in the extraction of potential candidates from a high number of left-right grasp pairs. In this work, we leverage three well-established measures from the dexterous manipulation literature to evaluate the bimanual grasp pairs and extract the reasonable ones.

%% file: 02_dataset.tex
\section{Dataset Details}
\label{sec:data_set}
In this section, details about the proposed DA$^2$ dataset are presented. In the first part, we describe some basic information about our data source and the format of dual-arm grasp pairs. In the following parts, we illustrate our pipeline to generate grasp pairs. Our code will be made publicly available via ~\url{https://sites.google.com/view/da2dataset/code}.

\subsection{Preliminary}
Our goal is to avoid labeling through complex and time-consuming manual setups. Since ACRONYM~\cite{eppner2021acronym} has shown the effectiveness of simulated data which has been successfully deployed in several works~\cite{yang2021reactive, wang2022goal, sundermeyer2021contact}, we also omit real data and use a simulation environment instead. We leverage the same source as in Eppner et al.~\cite{eppner2021acronym}, namely ShapeNetSem~\cite{savva2015semantically}, to obtain extended semantic information. Unlike the scaling policy used by them, we explicitly choose a factor to rescale the meshes so that the bounding box is scaled in the interval from 60 to 100~cm to represent large objects. Given the workspace of typical dual-arm robots, objects larger than 100~cm are not considered.

Each pair of grasp poses in our dataset is given by ${{\cal G}}=\left\{{{{\cal G}}}_{i}|i=1,2\right\}$, where ${{\cal G}_{i}} \in SE(3)$ is the 6-DOF pose of the left and respectively right end-effector pre-grasps with respect to the object frame.Our dataset is built for parallel-jaw grippers, and specifically, we use Robotiq 2F-85~\cite{WinNT} as the gripper model.
All pre-grasps in the dataset are created with~\SI{0.085}{\m} gripper width to match the Robotiq 2F-85 hardware.

\subsection{Block Antipodal Sampling}
\label{anti}
Antipodal sampling is a frequently used technique for grasp pose generation, which has been used in several previous works~\cite{mahler2017dex,eppner2021acronym,liang2019pointnetgpd}. Antipodal sampling constitutes a typical method~\cite{eppner2019billion,eppner2021acronym} to create a set of diverse grasps for small objects with the range of up to \SI{100}{k}~grasps per object. Given an object mesh, this scheme first samples an arbitrary point on the mesh surface as the initial contact point together with a line within a range around the surface normal. The sampling threshold $\gamma = \tan(\alpha/2)$ with the cone angle $\alpha$ restricts the range at which rays can be emitted. A second point is found as the intersection of both mesh and line. ${{\cal G}}_{i}$ is then derived by taking the center point between two contacts and a randomly sampled rotation around the line. While this method generates a set of grasp poses, some issues exist that hinder its direct application for large objects. Firstly, it is not guaranteed to find a line segment whose length between two contact points is {shorter} than the maximum width of the gripper fingers. Secondly, the density of sampled surface points for large objects is significantly sparser than that for smaller ones. Thus, more iterations are required to create a sufficient number of feasible grasp candidates for one end-effector.

To conquer such issues, we propose an incremental sampling scheme called \textit{Block Antipodal Sampling}. Assuming our goal is to obtain $g$ grasp pairs, we firstly crop meshes by dividing the bounding box of the original object into $i \times i \times i$ blocks, and then we can sample $g/i^3$ grasps in each block in parallel as illustrated in Fig.~\ref{fig:block_and_MSV}.
We iteratively check whether or not the generated grasps collide with the object mesh and produce new grasps until we either found enough poses or the number of iterations has reached its maximum. Once the feasible grasps for a single end-effector are generated, we can obtain $C_{g}^{2}$ grasp pairs by random combination. Note that this step will generate millions of feasible grasp pairs if no additional evaluation metrics are applied. In Section~\ref{quality}, we prune the grasp pairs through force closure inspection and set the maximum number of grasp pairs per object to 2001.

\subsection{Grasp Dexterity Labeling}
\label{quality}
\textit{Grasp dexterity} defines the ability of a grasp to achieve one or more useful secondary objectives while satisfying the kinematic relationship (between joint and Cartesian spaces) as the primary objective. To reduce unnecessary costs, we need to select feasible grasp pairs with various dexterity to form a balanced dataset from the generated candidates. Different from the single-arm optimal grasp generation, evaluating the grasp dexterity and labeling the grasp pair for the DAH system are much more involved in terms of grasp stability analysis.
We propose to evaluate the grasp quality for the DAH system from three aspects, {namely grasp stability, minimum singular value, and force minimization}. We detail these aspects hereafter:

\noindent
{\textbf{Grasp stability:}} The generated grasp pair should satisfy the force-closure condition, which means that with this grasp, the DAH system can generate wrenches to the coordinately manipulated object to resist random external disturbances in any direction. PCWF is used as the contact model to characterize the force transmission between each finger of each end-effector and the object.
A grasp pair for the DAH can be represented by a set of contact point pairs ${\cal P} =\{ {\left( {{{\mathbf p}_{l,1}},{{\mathbf p}_{r,1}}} \right),\left( {{{\mathbf p}_{l,2}},{{\mathbf p}_{r,2}}} \right)|{{\mathbf p}_{l,i}},{{\mathbf p}_{r,i}}} \in {\mathbb{R}^3},i = 1,2 \}$, where ${\mathbf p}_{l/r,i}$ denotes the \emph{i}th contact point of the left/right arm on the object surface, respectively. To avoid slippage of the fingertips, the contact force ${\mathbf f}_{l/r,i} \in {\mathbb{R}^3}$ applied by each finger of the left/right hand at each contact point is constrained to lie in the respective friction cone $\left\{ {\left( {{{\mathbf c}_{l,i}},\mu } \right) \cup \left( {{{\mathbf c}_{r,i}},\mu } \right)} \right\}$ where ${{\mathbf c}_{l/r,i}}$ denotes the friction cone axis at the \emph{i}th contact point of the left/right arm and $\mu$ represents the friction coefficient of the object surface. Inspired by~\cite{liu2021synthesizing}, the constraints imposed by the force closure on a given grasp configuration can be relaxed as
\begin{subequations}
    \begin{align}
        {\mathbf G}{{\mathbf G}^T} &\succeq \varepsilon {{\mathbf I}_{6 \times 6}}, \label{eq_2c1}\\
        {\mathbf G}{\mathbf f} & = 0, \label{eq_2c2}\\
        {\left\| {{\mathbf G}{\mathbf c}} \right\|_2} &={\left\| { - \frac{{{\mathbf G}{{\mathbf f}_t}}}{{{{\left\| {{{\mathbf f}_n}} \right\|}_2}}}} \right\|_2}= \omega, \label{eq_2c3}
    \end{align}
    \label{eq_2c}%
\end{subequations}%
    where  $\mathbf f ={\mathbf f}_t+{\mathbf f}_n = [ {{\mathbf f}_{l,1}^T,{\mathbf f}_{l,2}^T,{\mathbf f}_{r,1}^T,{\mathbf f}_{r,2}^T} ] \in {{\mathbb{R}}^{12}}$ is the collective contact forces exerted by the DAH system, which is unknown here. The parts ${\mathbf f}_t$ and ${\mathbf f}_n$ respectively denote the tangent and normal components and ${\mathbf c} ={[ {c_{l,1}^T,c_{l,2}^T,c_{r,1}^T,c_{r,2}^T} ]^T} \in {{\mathbb{R}}^{12}}$ is the collective vector of all the friction cone axes and can be easily obtained through typical shape representations of the objects. ${\mathbf G} = [ {{{\mathbf G}_{l,1}},{{\mathbf G}_{l,2}},{{\mathbf G}_{r,1}},{{\mathbf G}_{r,2}}} ]$ with ${{\mathbf G}_{l/r,i}} = {[ {{{\mathbf I}_{3 \times 3}},{{( {{{\lfloor {{{\mathbf p}_{l/r,i}}} \rfloor }_ \times }} )}^T}} ]^T}$ denotes the grasp matrix, ${{{\left\lfloor {{{\mathbf p}_{l/r,i}}} \right\rfloor }_ \times }}$ is the skew symmetric matrix of ${{{\mathbf p}_{l/r,i}}}$.
To avoid solving the unknown forces $\mathbf f$
we chose a small positive constant $\varepsilon > 0$ that guarantees the positive-definiteness of ${\mathbf G}{{\mathbf G}^T}$ and $\omega$ is introduced to reflect the angle between the friction cone axis and the applied force, which can be viewed as a measure of the grasp stability.
A small $\omega$ value indicates that the actual force has a higher probability of lying within the friction cones.
 We then arrange the grasp pairs into three parts according to $\omega$ from smallest to largest, with an equal number of components in each part, and select up to 667 best components in each part to form the final set of grasping pairs for an object.
 
 \remark{According to the grasp theory, the effectiveness of using $\omega$ is supported by that: ${{\mathbf G}{\mathbf c}}/4$ is guaranteed to be inside the grasp wrench space and force closure requires the origin of the wrench space to be inside the grasp wrench space. Hence, the distance between ${{\mathbf G}{\mathbf c}}/4$ and the origin, namely $\|{{\mathbf G}{\mathbf c}}/4\|_2$ or equivalently $\|{{\mathbf G}{\mathbf c}}\|_2$, can in some sense reflect how far the grasp is from having force closure, and it is also related to the inclination angle of the contact forces satisfying \eqref{eq_2c2}~\cite{zheng2006enhanced}.}

\noindent
{\textbf{Minimum singular value:}} The minimum singular value (MSV) {of the grasp matrix $\mathbf {G}$ should be maximized to guarantee that the DAH-object system with the grasp configuration always has the capability of withstanding external wrenches in all directions}. The MSV employed here is to measure the distance to the singular configuration of bimanual grasp, with which grasp pairs that are prone to singularity-related task failures can be filtered out. Previous considerations~\cite{shimoga1996robot} have shown that MSV varies significantly near singular configurations.
Using singularity value decomposition of the grasp matrix ${\mathbf G}$, we can obtain 
\begin{align}
\label{MSV}
\mathbf G = {\mathbf U}{\mathbf \Sigma} {{\mathbf V}^T},
\end{align}
where $\mathbf U \in {\mathbb{R}^{6 \times 6}}$ and ${\mathbf V} \in {\mathbb{R}^{12 \times 12}}$ are two real orthogonal matrices.  $\mathbf{\Sigma}$ is a $6 \times 12$ rectangular diagonal matrix with non-negative singular values $\{{\sigma _1},{\sigma _2},..., {\sigma _m}\}$ of $\mathbf G$ on the diagonal which are arranged in a descending order and $m \le 6$ with
\begin{align}
    \label{singular_value}
    {\sigma _1} \ge {\sigma _2} \ge ... \ge {\sigma _m},
\end{align}
where $\sigma _m$ denotes the MSV of the grasp pair candidate. 
We discover a regular pattern that the candidate grasp pair with maximum MSV tends to be symmetrically distributed around the centroid of an object (see Fig.~\ref{fig:block_and_MSV} for an example). Grasp pairs with ideal MSV will balance the grasping force. 

\noindent
{\textbf{Force minimization:}} To further evaluate the grasp pair candidates in terms of energy consumption, the grasp configuration with least force requirements on the DAH system is typically preferred. Here we introduce the concept of force ellipsoid~ {\cite{shimoga1996robot}} to depict the relationship between the grasp matrix and the force requirement, which is expressed as
\begin{align}
    \label{force_ellipsoid}
    {\mathbf f}_{ext}^T{\left( {{\mathbf G}{{\mathbf G}^T}} \right)^{ - 1}}{{\mathbf f}_{ext}} \le 1,
\end{align}
where the resultant force ${{\mathbf f}_{ext}}$ applied at the object frame {is generated by the normalized contact forces $\mathbf f$ with a given grasp configuration ${{\mathbf G}}$ and they} are related as ${{\mathbf G}}{{\mathbf f}} = {{\mathbf f}_{ext}}$. {It is worth mentioning that \eqref{eq_2c2} is the homogeneous equation of ${{\mathbf G}}{{\mathbf f}} = {{\mathbf f}_{ext}}$.} For a given ${{\mathbf f}_{ext}}$ balancing the desired object dynamics, a grasp configuration with least forces ${{\mathbf f}}$ is desirable. Since the major axis of this force ellipsoid denotes the direction in which the DAH system can apply {the largest} forces to the object for a given ${\mathbf f}$, we use the angle $\theta _{\mathbf{G}}$ between the direction of gravity and the major axis of the force ellipsoid for a grasp configuration as a new metric to measure the transformation efficiency. We note
\begin{align}
    \label{theta_g}
    \theta _{\mathbf G} = \arccos \left( {\frac{{{{ {\mathbf v}}_{major}} {\cdot {{ {\mathbf v}}_{gravity}}}}}{{\left\| {{{ {\mathbf v}}_{major}}} \right\|\left\| {{{ {\mathbf v}}_{gravity}}} \right\|}}} \right),
\end{align}
where ${{\mathbf v}}_{major}$ denotes the direction vector of the major axis of the ellipsoid, which coincides with the normalized eigenvector corresponding to the smallest eigenvalue $\lambda$ of ${{\mathbf G}{{\mathbf G}^T}}$ considering that the length of the major axis is
$1/\sqrt{\lambda}$,
and ${ {\mathbf v}}_{gravity}$ is the directional vector of gravity. Noting that ${\lambda}$ and $\sigma _m$ are related by $\sigma _m=\sqrt{\lambda}$. 
    \begin{remark}
    It is worth mentioning that the preferred force direction can be replaced according to specific task requirements. We assume that the object's gravity is the major force to be resisted in, e.g., pick-place or handover tasks. 
    \end{remark}

%% file: 04_graspnet.tex
\begin{figure}[t]
    \centering
    \includegraphics[width=0.95\linewidth]{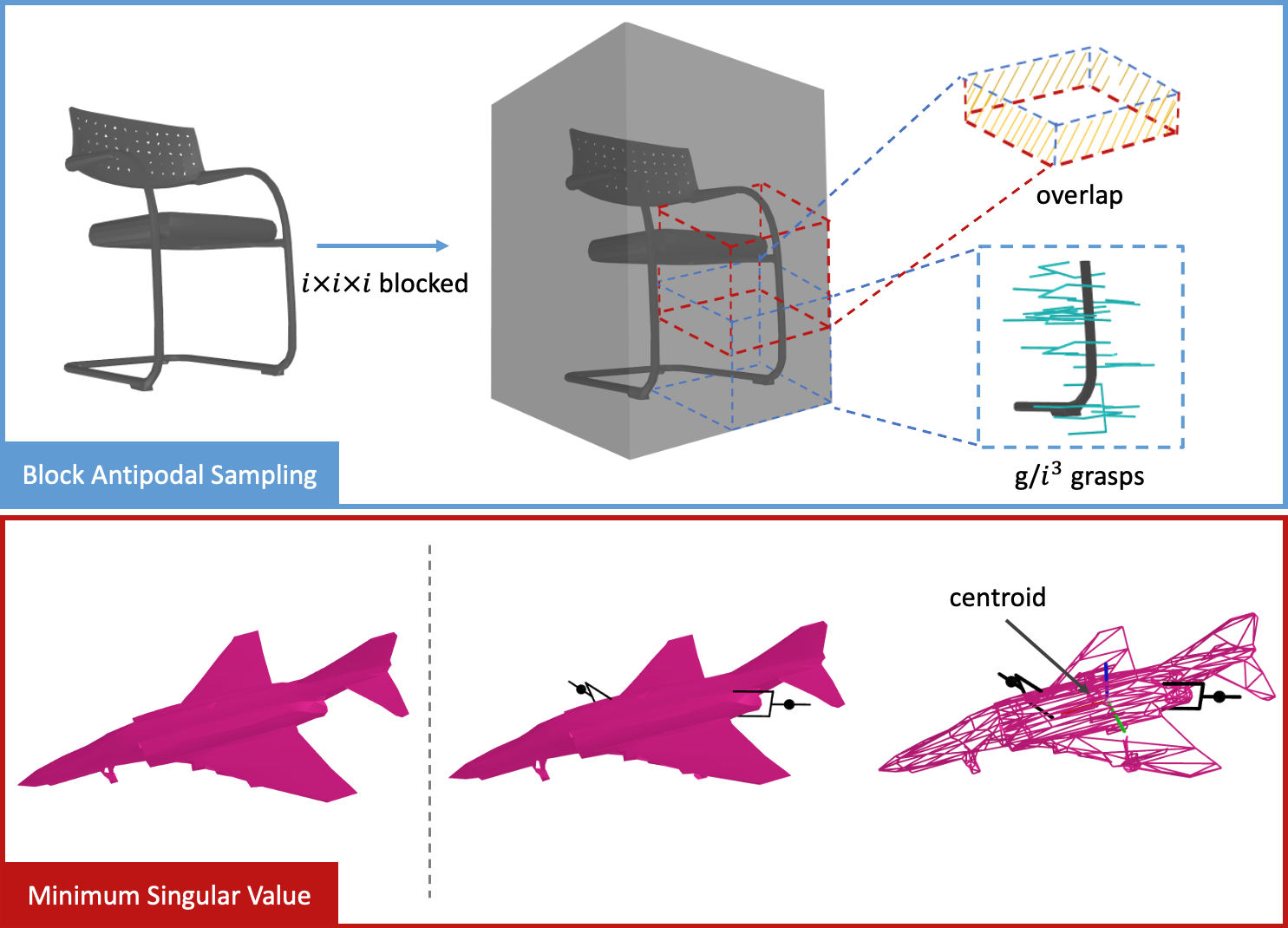}
    \caption{(Top) We perform antipodal sampling in each block, aligning adjacent blocks (red and blue) with a little overlap to prevent some potential grasps across both blocks from vanishing. (Bottom) An example shows the effectiveness of maximum MSV.}
    \label{fig:block_and_MSV}
    \vspace{-3mm}
\end{figure}
\begin{figure*}[t]
    \centering
    \includegraphics[width=0.95\linewidth]{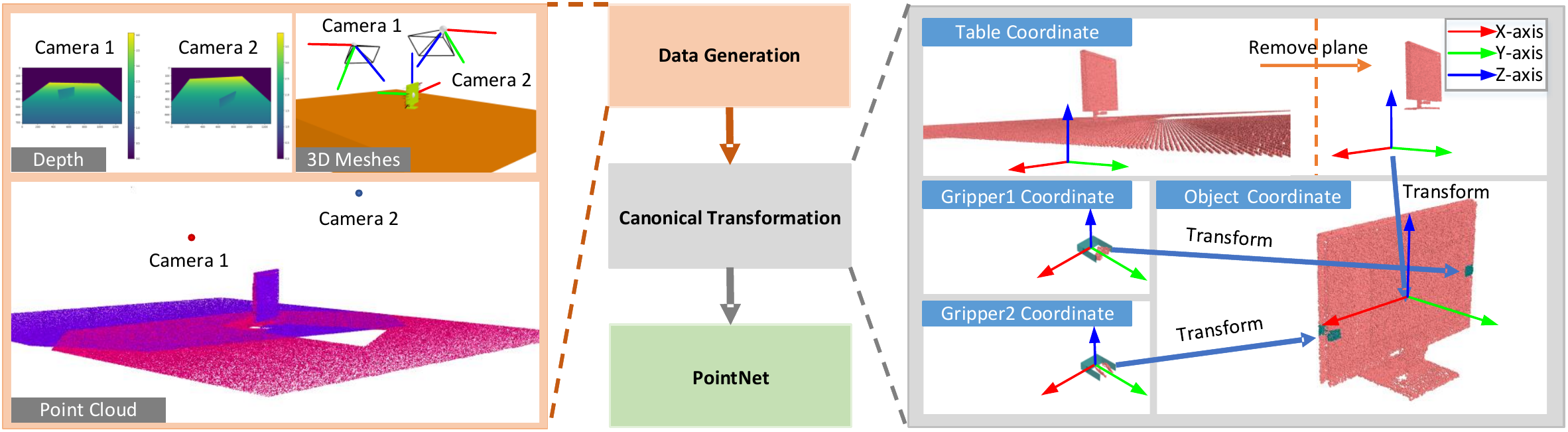}
    \caption{{Dual-PointNetGPD pipeline. Data generation module (left) generates depth images using pyrender, and the point cloud is obtained with the intrinsics of the virtual camera. After this step, points in the gripper closing volume are transformed to the object coordinates as presented in the canonical transformation module (right). Finally, transformed points are sent to PointNet to perform a classification task supervised by ground truth data.}}
    \label{fig:pipeline}
    \vspace{-5mm}
\end{figure*}
\section{Grasp Quality Evaluation Baseline}
To validate the performance of the generated dataset for optimal pair candidates, we propose an end-to-end baseline module called Dual-PointNetGPD based on PointNetGPD~\cite{liang2019pointnetgpd}. The main idea of~\cite{liang2019pointnetgpd} is to build a map between local features in points and the qualities of grasps through PointNet~\cite{qi2017pointnet}. As our grasp measures are based on grasp matrix analysis and the coordinates of the contact points, this basic mapping idea is also applicable to our case when evaluating the grasp quality. Compared with the non-dual case, Dual-PointNetGPD modifies the data source, coordinate representation, and quality labels to adapt to  specific requirement of the DAH system. The pipeline is shown in Fig.~\ref{fig:pipeline}.

\subsection{Train-val Generation}
Our train-val data are different from the original PointNetGPD. While PointNetGPD uses real scanning data from the YCB dataset~\cite{calli2015ycb} to train simulated grasps, we use the meshes in our simulated dataset. Following the instructions in Contact-GraspNet~\cite{sundermeyer2021contact}, we render point clouds by virtual depth cameras in different views. More concretely, we construct a scene with a virtual table being the support surface. Since the force minimization requirement we use in the dataset is highly relevant to the direction of gravity, we search for a stable pose $\mathbf{T}_{obj\leftarrow table}$ where the inner product of the $Z$-axis normal vector and the gravity is the smallest. Such a pose indicates that the object is ``standing" on the table, thus guaranteeing that all objects share consistent coordinate representation. The rotation angle around the $Z$-axis is sampled uniformly. We then position the camera on a hemisphere centered around the table midpoint where the optical axis points towards the table center. We choose a sphere radius between 1.5 and 2.0~m and render point clouds from this viewpoint. Additionally, we use two cameras which are antipodal with some perturbations, as the point cloud rendered by only one camera cannot cover the positions for arbitrary grasp pairs satisfactorily. We train Dual-PointNetGPD on the simulated scenes and validate it with real point cloud data to verify that models trained on DA$^2$ have the potential to be transferred to the practical applications.

\subsection{{Canonical Transformation}}
PointNetGPD crops and transforms the point cloud within the gripper into the gripper's local coordinates for each grasp candidate and feeds the points into the neural network.
This process eliminates ambiguity caused by using different camera coordinate frames.
The situation is entirely different for dual-arm grasp pose estimation.
It is hard to define which gripper is the optimal reference candidate.
At the same time, it is beneficial to choose a fixed global frame as the reference frame so that the coordinates of the inner points will not change with respect to each object to improve feature embedding and classification performance. This facilitates the convergence of PointNet~\cite{qi2017pointnet} as the backbone used in PointNetGPD. While it is intricate to set a global coordinate system for all data{, considering that the dexterity measures for labeling the bimanual grasp pairs are all based on the algebraic properties of the grasp matrix and $\mathbf{G}$ is naturally expressed with respect to the object frame, the object frame is thus a reasonable candidate as the reference frame since it can be considered as a relatively fixed frame for all grasp pairs on the object surfaces regarding the metrics and is available directly from meshes}. Thus we transform all inner points of the object initially from the camera frame $\mathbf{P}_{cam}$ to the object frame $\mathbf{P}_{obj}$ whose origin coincides with the centroid. Since the table frame can be treated as the world frame in the rendered scene, i.e., $\mathbf{T}_{table}=\mathbf{I}_{4 \times 4}$, we can obtain $\mathbf{P}_{obj}$ by using the transformation between camera and object frame:
\begin{equation}
	\begin{gathered}
       \mathbf{T}_{obj\leftarrow cam} = \mathbf{T}_{obj\leftarrow table}  \mathbf{T}_{table\leftarrow cam} \\
       	\mathbf{P}_{obj} =\mathbf{T}_{obj \leftarrow cam} \mathbf{P}_{cam}.
	\end{gathered}
\end{equation}

For practical implementation, we consider the average of the coordinates of all points on the object $\mathbf{t}$ as an equivalent replacement for the centroid.
The rotation part $\mathbf{R}$ of the stable pose can be calculated as:
\begin{equation}
	\begin{aligned}
		&\mathbf{R}=\left[\mathbf{n}_{x}, \mathbf{n}_{y}, \mathbf{n}_{z} \right] \\	
		&\mathbf{n}_{x}=\frac{1}{\sqrt{\mathbf{n}_{z^1}^{2}+\mathbf{n}_{z^2}^{2}}}\left[\mathbf{n}_{z^2},-\mathbf{n}_{z^1},0\right]^{\text{T}} \\ 
		&\mathbf{n}_{y}=\mathbf{n}_{z} \times \mathbf{n}_{x},
	\end{aligned}
\end{equation}

\noindent
where $\mathbf{n}_{z}$ denotes the unit normal vector of the support surface, which is usually a visible plane in the camera frame. Thus, ${{\mathbf T}_{obj \leftarrow cam}} = {\left[ {{\mathbf R},{\mathbf t};{\mathbf 0},1} \right]^{ - 1}}$.

After obtaining $\mathbf{P}_{obj}$ of both simulation training and the real test, we input them into a PointNet architecture to filter out the optimal grasp pair with a classifier.


\begin{figure}[t]
\centering
\subfloat[Number of grasp pairs vs. grasp stability under different $\gamma$.]{\label{distribution1}\includegraphics[width=0.95\columnwidth]{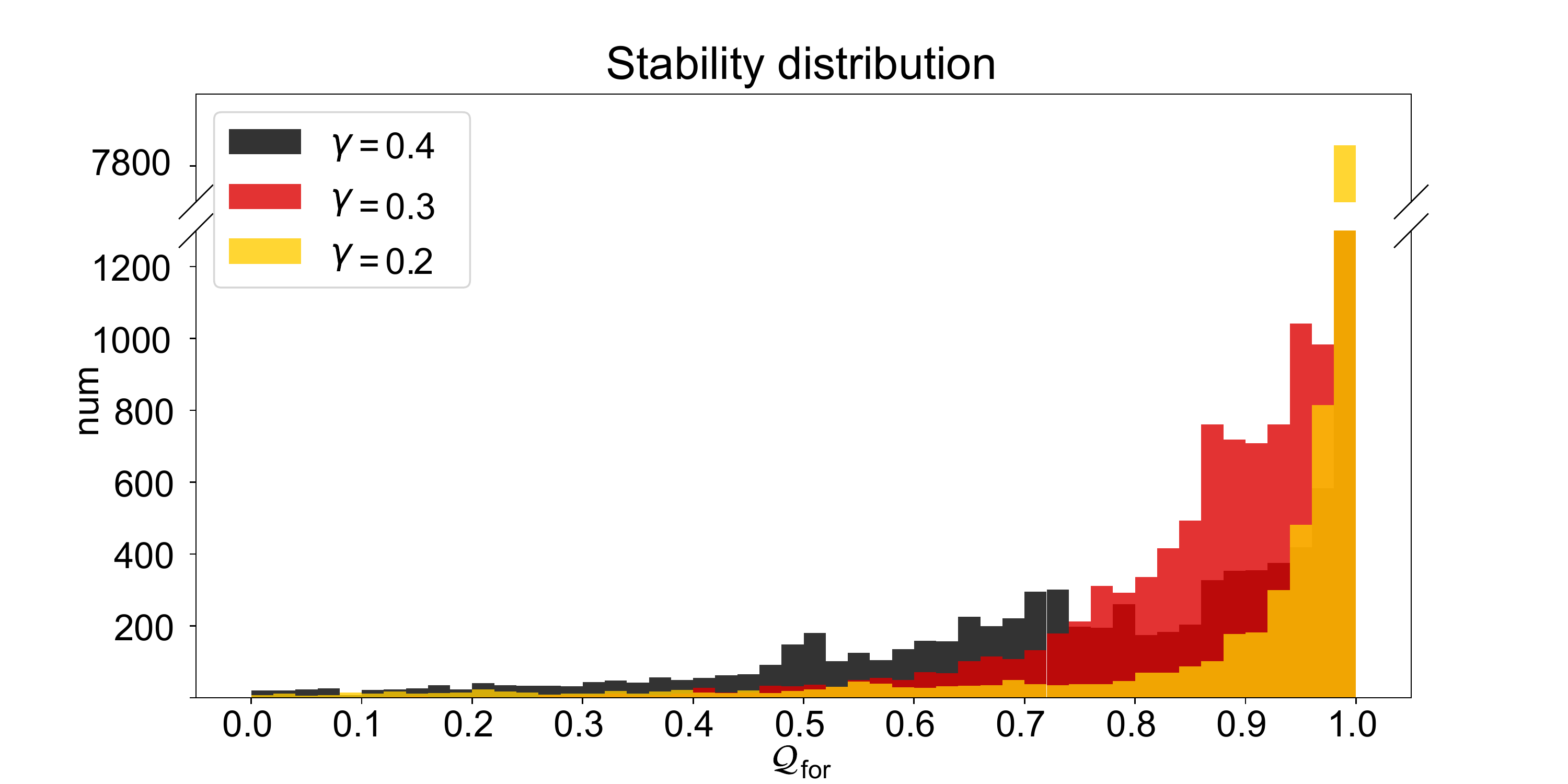}}

\subfloat[Number of grasp pairs vs. combined measures for simulation.]{\label{distribution2}\includegraphics[width=0.95\columnwidth]{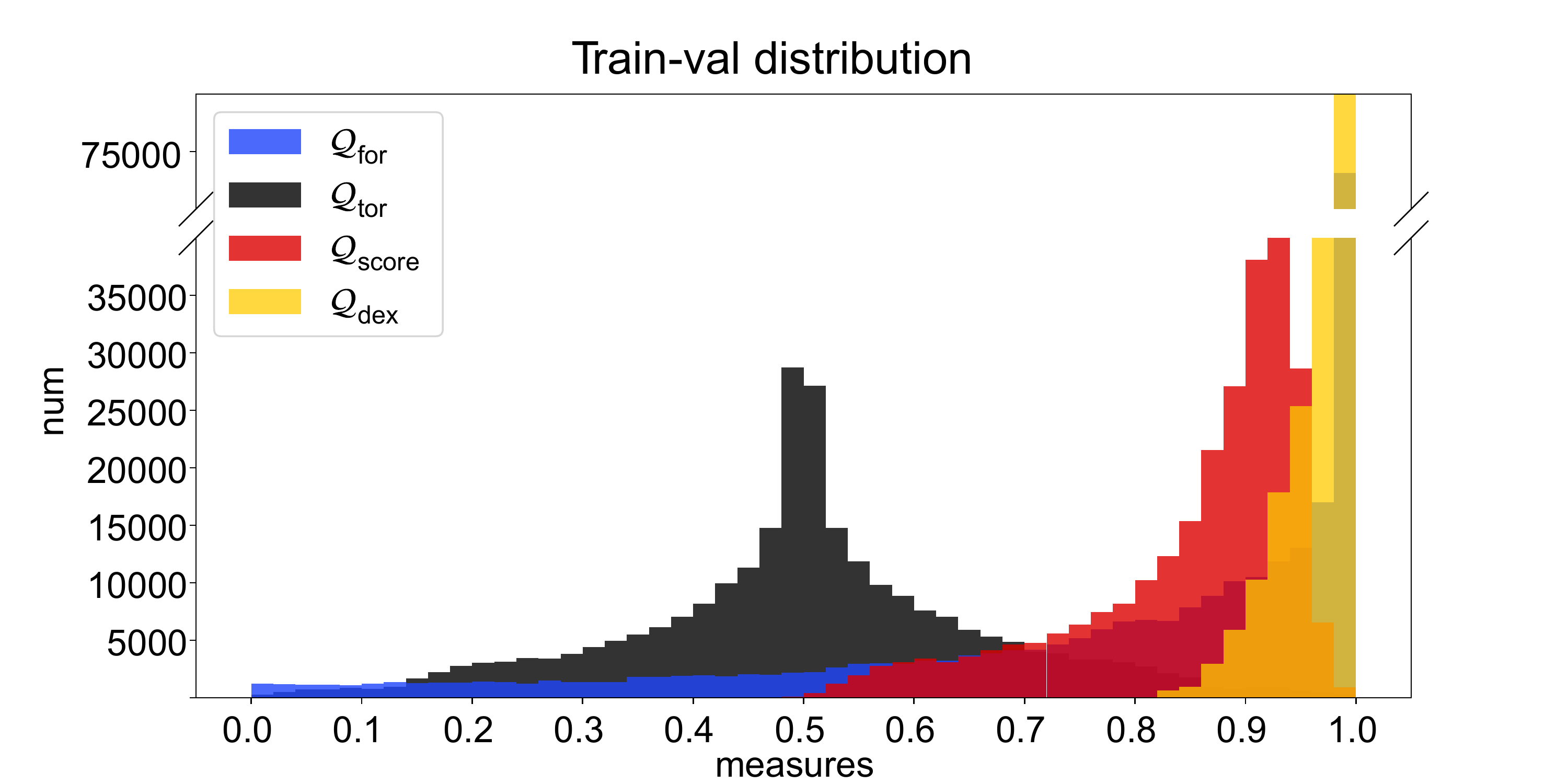}}
\DeclareGraphicsExtensions.
\caption{{Data distribution.}}
\label{fig:distribution}
\end{figure}
\vspace{-5mm}

\subsection{Combined Measures}
Considering multiple characteristics will influence the bimanual grasp performance. We combine the three measures mentioned in Section~\ref{quality} to evaluate the grasp quality of the pair candidates. The approach to balance the two measures in PointNetGPD is empirical. However, if the label is not normalized, gradients may oscillate and convergence speed {would probably} be slow.
Different from PointNetGPD, we decide to normalize each measure in DA$^2$ to the range $\left(0,1\right]$. We adjust weights applied to the three measures such that the network is capable to learn and evaluate grasp pairs of varying degrees. We note down the score
\begin{equation}
\label{metric}
\mathcal{Q}_{score}^k = \alpha\mathcal{Q}_{for}^k+\beta\mathcal{Q}_{dex}^k+\gamma\mathcal{Q}_{tor}^k,
\end{equation}
\begin{small}
\begin{equation}
	\mathcal{Q}_{for}^k=1-\omega^k,\ \mathcal{Q}_{dex}^k=\sigma_{m}^k/\max\sigma,\ \mathcal{Q}_{tor}^k=\cos \theta _{\mathbf G}^k,
\end{equation}
\end{small}

\noindent
where $\alpha+\beta+\gamma=1$, and $k$ represents the $k$-th grasp pair which belongs to one object. And $\sigma$ is the set of MSV of all objects in the training dataset, where $\sigma=\left\{\sigma_m|m=1,...,M\right\}$, $\sigma_m=\left\{\sigma_m^k|k=1,...,n\right\}$.

%% file: 05_results.tex
\section{Experiments}
\label{sec:baselines}
We believe DA$^2$ dataset can serve as a fair comparison benchmark for dual-arm optimal grasping, as it contains various labels and sufficient grasp data. In this section, we illustrate the evaluation of Dual-PointNetGPD that is trained on DA$^2$ following the evaluation protocol of previous works~\cite{sundermeyer2021contact, liang2019pointnetgpd}. We quantify the classification ability of the network on simulated data with ground truth data of every measure. In real tests, we evaluate the model with a robotic hardware experiment and focus on the grasping success rate by conducting several robotic grasping and R2H handover experiments. These serve as a basis to validate the effectiveness of our model in typical dual-arm coordinated tasks.

\subsection{DA\texorpdfstring{$^2$}{Lg} Exploration}
\label{Exploration}
{$i$ and $\gamma$ mentioned in Section III-B are the hyperparameters to be set for dataset generation. $i$ can be dynamically adapted, depending on the size of objects and the performance of the workstation. In this work, $i$ is fixed and set to be $3$, since the sizes of bounding boxes lie within a relatively narrow interval and are evenly distributed.}
As mentioned in Section~\ref{quality}, $\omega$ is the primary factor when we prune the extensive data. It is critical to ensure that $\omega$ is in a wide range. We examine the effect of different sampling factors $\gamma$ on $\omega$ during the dataset generation phase. We evaluated $\mathcal{Q}_{for}$ for 10k grasp pairs under different $\gamma = 0.2, 0.3, 0.4$ respectively. Fig.~\ref{distribution1} shows the stability distribution result. As $\gamma$ increases, more grasp pairs with lower $\mathcal{Q}_{for}$ appear, which indicates that DA$^2$ is covering a wider variety of cases. We choose $\gamma=0.4$ to obtain a balanced dataset in our final dataset creation.

\subsection{Simulation Details}
\label{Simulation}
\noindent
\textbf{Implementation details.} Following PointNetGPD, we chose a 3-class classification task. For the training aspect, we used an NVIDIA GeForce 1080Ti GPU and the Adam optimizer with an initial learning rate of 5e-3 with rate decay to half every 60 epochs. {For} classification {thresholds}, we chose 0.85 and 0.92, for worst and best respectively to balance the number of grasps in each bin (see Fig.~\ref{distribution2}). The weighting parameters in~\eqref{metric} are chosen as $\left(\alpha, \beta, \gamma\right)=(0.4, 0.5, 0.1)$ {such that the network focuses more on the bimanual grasp stability and simultaneously avoids the grasp singularity to a large extent}.

\noindent
\textbf{Simulation results.} As the first attempt to deal with dual-arm grasping, our method serves as a baseline, and here we report the classification accuracy as the evaluation metric. Specifically, we randomly rendered 380 scenes, including 248'102 grasp pairs for training. After this, we chose 19 scenes, 13'739 pairs, for validation. The variation of the accuracy of considered grasp pairs during training is demonstrated in Fig.~\ref{figure1}. After 250 epochs (model-250), the accuracy was stable at 0.72 with the reduction of the negative log likelihood loss. Finally, we tested model-250 in the real world.

\subsection{Real World Experiment}
\label{Real}
\noindent
\textbf{Experimental Setup.} We chose the dual-arm Kinova Gen3 manipulators with self-designed grippers. The visual sensors we used were two synchronized and calibrated Kinect Azure cameras. Additionally, we set up a table supporting the objects to guarantee that they are in the workspace of the manipulators.

\noindent
\textbf{Method Pipeline.}
Following the procedures in~\cite{liang2019pointnetgpd}, we chose GPG~\cite{ten2018using} to heuristically generate single grasps. Then, the single grasps were randomly combined to form grasp pair candidates, which were sent to model-250 to extract the optimal pair. As the classification results of top-k pairs are the same, we sorted the candidates by the probability outputs. Once the optimal one (i.e. the one with highest probability) was generated, the two manipulators were controlled by using the formulation of the extended cooperative task space~\cite{park2015extended} so that the end-effectors were aligned with the desired optimal grasp pair. Dynamic system~\cite{mirrazavi2018unified} was used to plan the relative and absolute trajectories for the DAH system. 

\noindent
\textbf{Experimental results.} As shown in Fig.~\ref{fig:hardware}, we selected seven objects and tested each object for five rounds with random initial orientations. If no feasible grasp pair is generated or if the object drops during R2H handover, we mark this attempt as failed. Table.~\ref{tab:single_grasp} shows the success rate for large object grasping using Dual-PointNetGPD. For keyboard grasping, we used an upholder to lift the keyboard so that the grippers could grasp it without collision. The tripod and those having shafts were easy to grasp. For the yellow chair, its geometric feature is difficult for GPG to find any proper grasps. Finding the optimal grasp pair for the box was successfully done but grasping sometimes failed due to the large box width which was difficult for the gripper to fit it in. In addition to these, some failures happened during handover even though the robot was able to find feasible grasping positions. We observed that in these cases, the contact surfaces were more slippery. Although the used grasp stability measure is independent of the friction coefficient, the domain gap can cause performance degradation and we observed that object weight has a significant influence on the grasping success. 
{It can be concluded from the experimental results that the employed grasp sampler, the capability of the DAH system and the reality gap all have an effect on the final success rate in practical implementations besides the network that is trained with the proposed dataset. We believe that this can inspire future research along these directions.}

%% file: 06_conclusion.tex
\begin{figure}[t]
	\sf
	\centering
	\normalsize
	\includegraphics[width=\linewidth]{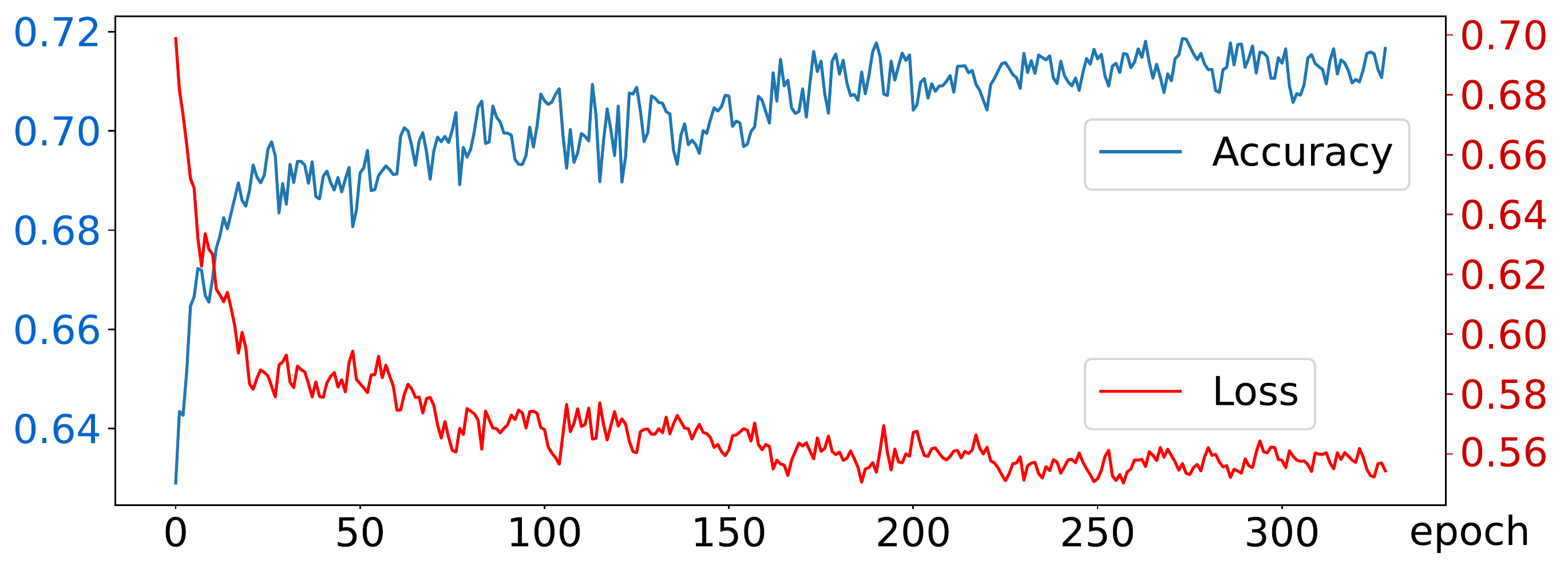}
	\caption{The variation of classification accuracy and loss on the validation set for different epochs during training.}
	\label{figure1}
\end{figure} 
\begin{table}[t]
\centering
\caption{Success Rate of Dual-Arm Large Object Grasping}
\begin{tabular}{m{0.66cm} m{0.66cm} m{0.66cm} m{0.66cm} m{0.66cm} m{0.66cm} m{0.66cm} m{0.66cm}}
\toprule
Avg. & Tripod & \begin{tabular}[c]{@{}c@{}}Yellow \\ stool\end{tabular} & \begin{tabular}[c]{@{}c@{}}Green\\ stool\end{tabular} & Box & Dustpan & Gray stool & \begin{tabular}[c]{@{}c@{}}Key- \\ board\end{tabular} \\ \midrule
0.66 & 1.00 & 0.00 & 0.60 & 0.80 & 0.80 & 0.80 & 0.60 \\ \bottomrule
\end{tabular}
\label{tab:single_grasp}
\end{table}

\section{Discussion}
\label{sec:limitations}

The research of vision-guided dual-arm optimal grasping is still in its infancy. It is also worth mentioning that the unique framework built with this project generalizes further to the DAH system with multi-fingered hands, for which the proposed dexterity labeling process can be adjusted through the extension of the grasp matrix. We believe that the dissemination of our dataset can enable the first of such investigations and ultimately pave the way to further dual-arm research in the future. Here we point out two promising aspects.

\begin{figure}[t]
    \centering
    \includegraphics[width=0.95\linewidth]{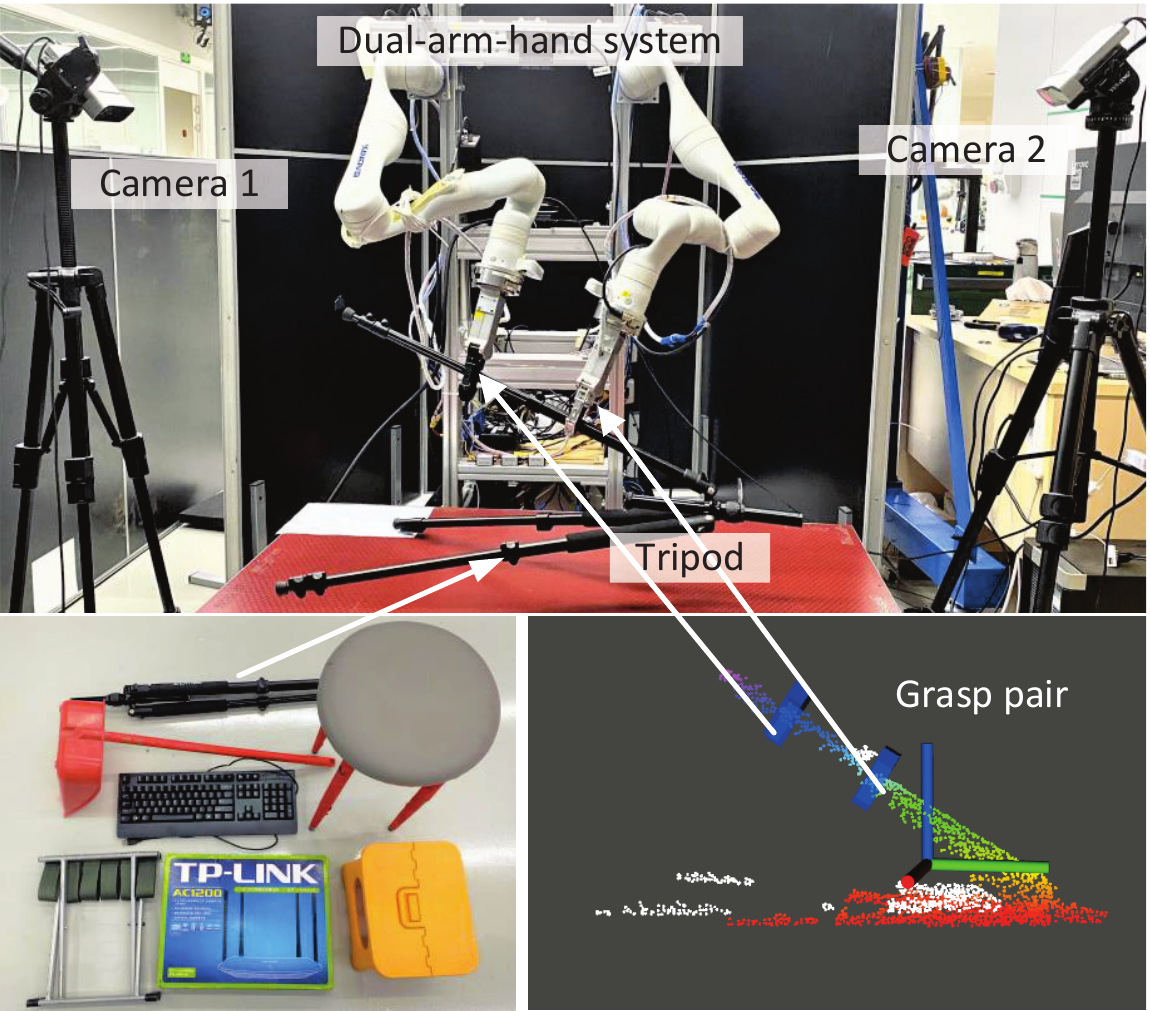}
    \caption{(Top) The agent actuator was picking up a tripod and then handed it to a human; (Bottom left) Various large objects; (Bottom right) Point cloud of the tripod and the optimal grasp pair candidate.}
    \label{fig:hardware}
\end{figure}

\noindent
\textbf{Synthetic vs. Real dataset.} Synthetic data enjoys low annotation cost due to the avoidance of label-intensive human annotations and it is manageable to arrange virtual scenarios.
 In line with previous works~\cite{mahler2017dex,eppner2021acronym}, our real experiment shows that training on synthetic data can serve as a reliable foundation for practical implementations.
However, a domain gap still exists and we believe that further adaptation and photorealistic rendering techniques~\cite{liu2022robotic} can be leveraged to further reduce this gap and continuously improve model performance.


\noindent
\textbf{Conventional vs. Unique Benchmark.}
Our experimental evaluation protocols allowed us {not only} to provide evidence for reliable classification compared to synthetic ground truth, but also to show {the effectiveness of} the proposed dual-arm grasping model for practical applications despite the synthetic-only training.
Using the conventional single-arm grasping benchmark enables the validation of the network classification accuracy, while the real application could be tested using the grasping success rate.
Given the underexplored area of dual-arm grasping with only limited considerations regarding stability and dexterity~\cite{roa2015grasp}, we think that case-aware metrics that include multi-modal factors such as grasp equilibria, kinematic considerations, tracking performance, as well as surface or object properties, might further advance a detailed understanding in this field.

\section{Conclusion}
\label{sec:conclusion}
This paper introduced the first large-scale dexterity-aware grasping dataset toward dual-arm grasping. Based on this novel dataset, an end-to-end pipeline named Dual-PointnetGPD was proposed as the critical component of the dual-arm grasping baseline and trained to evaluate the qualities of the grasp pair candidates effectively. Both simulation and real-robot experiments showed that our dataset provides distinct dexterity features, which can help the neural network to extract reasonable grasp pairs from numerous candidates in terms of dual-arm grasping dexterity. We hope both our dataset and baseline evaluation module can stimulate new algorithms in the dual-arm optimal grasping field.